\documentclass[runningheads]{llncs}
\usepackage{graphicx}
\usepackage{amsmath,amssymb} % define this before the line numbering.
\usepackage{color}

\usepackage[width=122mm,left=12mm,paperwidth=146mm,height=193mm,top=12mm,paperheight=217mm]{geometry}
\usepackage{hyperref}

\newcommand*\samethanks[1][\value{footnote}]{\footnotemark[#1]}
\newcommand{\bw}{\mathbf{w}}
\newcommand{\bx}{\mathbf{x}}
\newcommand{\bc}{\mathbf{c}}

\begin{document}
%\linenumbers
% \pagestyle{headings}
% \mainmatter
\def\ECCV18SubNumber{6}  % Insert your submission number here

\title{Constrained-size Tensorflow Models for YouTube-8M Video Understanding Challenge}
%\title{Restricted Size Deep Learning Model for Youtube8M Video Classification} % Replace with your title

\titlerunning{ECCV-18 submission ID \ECCV18SubNumber}

\authorrunning{ECCV-18 submission ID \ECCV18SubNumber}

\author{
Tianqi Liu\thanks{These two authors contributed equally}, Bo Liu\samethanks[1] 
\thanks{This author is with Kensho Technologies.}\\
\email{tianqi.terence.liu@gmail.com, bo.liu@kensho.com}
}

\institute{Paper ID \ECCV18SubNumber}

\maketitle

\begin{abstract}

This paper presents our 7th place solution to the second YouTube-8M video understanding competition which challenges participates to build a constrained-size model to classify millions of YouTube videos into thousands of classes. Our final model consists of four single models aggregated into one tensorflow graph. For each single model, we use the same network architecture as in the winning solution \cite{miech2017learnable} of the first YouTube-8M video understanding competition, namely Gated NetVLAD. We train the single models separately in tensorflow's default float32 precision, then replace weights with float16 precision and ensemble them in the evaluation and inference stages, achieving 48.5\% compression rate without loss of precision. Our best model achieved 88.324\% GAP on private leaderboard. The code is publicly available at \url{https://github.com/boliu61/youtube-8m}

\end{abstract}

\section{Introduction}

Enormous amount of video content is generated all over the world every day. As an important research topic in computer vision, video analysis has many applications such as recommendation, search, and ranking. Recently, video classification problem gained interest with broad range of applications such as emotion recognition \cite{ebrahimi2015recurrent}, human activity understanding \cite{caba2015activitynet}, and event detection \cite{xu2015discriminative}.

YouTube-8M dataset \cite{abu2016youtube} released by Google AI consists of over 6 million YouTube videos of 2.6 billion audio and visual features with 3,700+ of associated visual entities on average of 3.0 labels per video. Each video was decoded at 1 frame-per-second up to the first 360 seconds, after which features were extracted via pre-trained model. PCA and quantization were further applied to reduce dimensions and data size. Visual features of 1024 dimensions and audio features of 128 dimensions were extracted on each frame as input for downstream classifiers.

Following the first YouTube8M Kaggle competition, the second one is focused on developing compact models no greater than 1 GB uncompressed so that it can be applicable on user's personal mobile phones for personalized and privacy-preserving computation. Challenges in such competition include modeling correlations between labels, handling multiple sequential frame-level features, and efficient model compression. 

In the competition, Global Average Precision (GAP) at 20 is used as metric. For each video, the model should predict 20 most confident labels with associated confidence (probability). The list of $N$ tuples $\{video,label,confidence\}$ is sorted by confidence levels in descending order. GAP is then computed as:
\begin{equation}
GAP=\sum_{i=1}^Np(i)\Delta r(i)
\end{equation}
where $p(i)$ is the precision, and $r(i)$ is the recall.

Common approach for video analysis typically extract features from consecutive features followed by feature aggregation. Frame-level feature extraction can be achieved by applying pre-trained Convolutional Neural Networks (CNN) \cite{he2016deep,krizhevsky2012imagenet,simonyan2014very,szegedy2017inception}. Common methods for temporal frame feature aggregation include Bag-of-visual-words \cite{sivic2003video,csurka2004visual}, Fisher Vectors \cite{perronnin2007fisher}, Convolutional Neural Networks (CNN) \cite{Karpathy_2014_CVPR}, Gated Recurrent Unit (GRU) \cite{ballas2015delving}, Long Short-Term Memory (LSTM) \cite{yue2015beyond}, and Generalized Vector of Locally Aggregated Descriptors (NetVLAD) \cite{arandjelovic2016netvlad}. 

It is well-known that neural networks are memory intensive and deploying such models on mobile devices is difficult for its limited hardware resources. Several approaches were proposed to tackle such difficulty. A straight forward way is to apply tensor decomposition techniques \cite{zhang2001rank,kolda2009tensor,liu2017characterizing} to a pretrained CNN model \cite{denton2014exploiting}. Network Pruning removes low-weight connections on pretrained models \cite{han2015learning}, or gradually trains binary masks and weights until target sparsity is reached \cite{zhu2017prune}. Network quantization compresses network by reducing number of bits required to represent each weight via weight sharing \cite{han2015deep}, or vector quantization \cite{gong2014compressing}. Another way to get better performance for limited-size model is to use knowledge distillation \cite{hinton2015distilling,romero2014fitnets}. The idea behind it is to train student network (small size) to imitate the soft output of a larger teacher network or ensembles of networks.   

In Google cloud \& YouTube-8M video understanding challenge Kaggle competition, top participates \cite{miech2017learnable,wang2017monkeytyping,li2017temporal,chen2017aggregating,skalic2017deep} trained models such as Gated NetVLAD, GRU, and LSTM with Attention. To leverage the predictability of single models, they averaged checkpoints at different training steps and ensembled predicted probability scores by weighted average, bagging, or boosting. 

Our contribution in this paper is threefold: First, we explore size and performance of Gated NetVLAD under different sets of hyper-parameters (cluster size and hidden size). Second, we develop ensemble approach of multiple models in one tensorflow graph which avoids in-place change of graph. Third, we cast trained weights tensors from float32 to float16 data type in evaluation and inference stage which reduces approximately half the model size without sacrificing performance.

The rest of the paper is organized as follows. Section 2 presents our model architecture with compression and ensemble approaches. Section 3 reports experimental results, followed by conclusions in Section 4.

\section{Approach}
In this section, we first introduce Gated NetVLAD model architecture, then describe model compression approaches we have tried, followed by ensemble approaches we developed.
\subsection{Frame Level Models}\label{subsec:single_model}
\begin{figure}
\centering
\includegraphics[height=2.5cm]{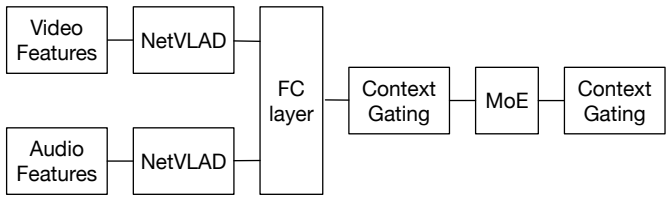}
\caption{Overview of our model architecture. FC denotes Fully Connected layer. MoE denotes Mixture of Experts.}
\label{fig:network_architecture}
\end{figure}

Our architecture for video classification is illustrated in Fig.~\ref{fig:network_architecture}

\subsubsection{NetVLAD}
\ \\

The NetVLAD \cite{arandjelovic2016netvlad} is a trainable generalized VLAD layer that captures information about the statistics of local descriptors over the image, i.e., the sum of residuals for each visual word. More specifically, let $i$th descriptor (video or audio feature) of a video be $x_i$, which can be assign to one of $K$ clusters with centroid $c_k$ for $k\in [1,K]$. The NetVLAD can be written as summation of residuals with soft assignment
\begin{equation}
V(j,k) = \sum_{i=1}^N\frac{e^{\bw^T_k\bx_i+b_k}}{\sum_{k'}e^{\bw^T_{k'}\bx_i+b_{k'}}}(x_i(j)-c_k(j))
\end{equation}
where $\{\bw_k\}$, $\{b_k\}$ and $\{\bc_k\}$ are sets of trainable parameters for each cluster $k$. Number of clusters (referred as cluster size) $K$ is a hyper-parameter we varies across different models. 

\subsubsection{Fully Connected layer}
\ \\

Our FC layer consists of two layers. First layer get input of concatenated video and audio VLAD layer, multiplied by weight matrix resulting in hidden size $H$, followed by batch normalization and ReLU activation. Second layer takes output of first layer as input, multiplied by weight matrix of shape $H\times 2H$ and added by a bias term.   

\subsubsection{Context Gating}
\ \\

Context Gating (CG) \cite{miech2017learnable} is a learnable non-linear unit aiming to model interdependencies among network activations by gating. Concretely, CG transform input vector $X\in \mathbb{R}^p$ to output vector $Y\in \mathbb{R}^p$ of same dimension via
\begin{equation}
Y = \sigma(WX+b)\circ X
\end{equation}
where $W\in \mathbb{R}^{n\times n}$ and $b\in \mathbb{R}^n$ are trainable parameters. $\sigma$ is element-wise sigmoid activation and $\circ$ is the element-wise multiplication. CG is known for being able to capture dependencies among features.

\subsubsection{Mixture of Experts}
\ \\

Mixture of Experts (MoE) \cite{jacobs1991adaptive} consists of two parts, gating and experts. The final predictions are sum of products of last layers from gating and experts. We use 5 mixtures of one hidden layer experts. Gate activations are multiplied to each expert for probability prediction. MoE is further followed by CG modeling dependencies among video vocabulary labels. 

\subsubsection{Training}
\ \\

We use two local GPUs, Nvidia 1080Ti and Nvidia 1060, and two Google Cloud Platform accounts with Tesla K80 GPUs to train single models. The training time is two to three days per model for 200k steps with batch size between 80 and 160.\\

The YouTube-8M Dataset is partitioned into three parts: Train, Validate and Test. Both Train and Validate data come with labels so they are effectively exchangeable. In order to maximize the number of training samples as well as to speed up the evaluation step, we randomly chose 60 out of the 3844 validate files as our validate set, and combine the other validate files with the official train dataset as our training set. We observe constant delta between our validate set and public leaderboard.

\subsection{Compression}
\subsubsection{Float16 Compression}
\ \\

Of the compression methods we have attempted, this is the only one that worked. The idea is to train all the models in the default float32 precision to achieve maximum score, then at evaluation/inference stage, cast all the inference weight tensors to float16 precision to cut model size in half while preserving prediction accuracy.\\

In actual implementation, we only cast 4 out of the 32 weight tensors in our model architecture. This is because the largest 4 tensors (\begin{verbatim}`tower/experts/weights',
`tower/gates/weights',
`tower/gating_prob_weights', 
`tower/hidden1_weights/hidden1_weights'\end{verbatim}) make up about 98\% of total model size. Modifying the other 28 small weight tensors does not worth the effort in our opinion, since float32 is the default data type in many tensorflow modules and functions, and we had to extend tensorflow's \texttt{core.Dense} class and \texttt{layers.fully\_connected} function to support float16 precision in order to cast those 4 tensors.\\

The average compression rate is 48.5\% across models, as shown in Table \ref{table:single_model}. Compared to the original float32 models, the float16 compression version perform equally well, with GAPs differing less than 0.0001, which is the level of randomness between different evaluation runs for the same float32 model.

\subsubsection{Gradual Binary Mask Pruning}
\ \\

We tried method introduced by \cite{zhu2017prune} as the tensorflow implementation is available on tensorflow's github repo. For every layer chosen to be pruned, the authors added a binary mask variable of the same shape to determine which elements participate in forward execution of the graph. They introduced gradual pruning algorithm updating binary weight masks along with weight in network training. They claimed to achieve high compression rate without losing significant accuracy.

However, we found two main difficulties of the method in our experiments:
\begin{enumerate}
\item The sparsity is not identical to compression rate. In the article, sparsity was referred as the ratio between number of non-zero elements of the pruned network and the original network, while after compression the sparse tensor has indices that takes additional storage. Although the authors considered bit-mask and CSR(C) sparse matrix representation, there are still two problems we could not solve easily: First, Tensorflow boolean variable takes one byte (not one bit) thus in real implementation the compression rate will be much lower. Second, for large tensors with huge indices range, row and column indices should have each element of type 32 or 64 bit integers, which takes a huge storage.

For example, if we use float32 to store a tensor of 1024x100000 and set sparsity as $75\%$. To make use of SparseTensor object in Tensorflow, for each non-zero element, we need to associate it with two 32 bit integers (row and column index). This results in only $25\%$ compression rate though sparsity is set to be $75\%$. Furthermore, it losses $75\%$ of non-zero elements and sacrifices accuracy with only $25\%$ compression rate. This is not as appealing as float16 compression approach.

\item At the time of competition deadline, the github repo \footnote{https://github.com/tensorflow/tensorflow/tree/\\
aa15692e54390cf3967d51bc60acf5f783df9c08/tensorflow/contrib/model\_pruning} for model pruning only contained the training part that associates tensor to be pruned with binary masks. The sparse tensor representation and bit mask were not implemented at the moment. We implemented it ourselves and found the compression rate not satisfiable. 

\end{enumerate}
After some trials and errors and comparison with float16 compression procedure, we decided not to pursue pruning for model compression.
 
\subsubsection{Quantization}
\ \\

Quantization \cite{han2015deep} can achieve $75\%$ compression rate by representing element in tensor as 8 bits unsigned integers rather than 32 bits float value. Tensorflow provides useful tools to quantize a pretrained graph. However, we found the output of graph to be frozen graph of pb format. We were able to convert meta, index, and checkpoint files to pb file, but not vice versa. Yet the competition requires that the submitted model must be loadable as a TensorFlow MetaGraph. We did not overcome this technical difficulty and went with our float16 compression method for its easy and elegant usage. 

\subsection{Ensemble}
\subsubsection{Checkpoint Average}
\ \\

For each single model, we first evaluate validation set using single checkpoints at some interval (say, every 10k steps) in order to know the approximate range of steps corresponding to higher scores for a particular model. Then we select a few ranges with varying starting and ending steps, and average all the checkpoints in that range. Finally we evaluate using these average points and pick the best one to represent this single model in the ensemble. On average, the averaged checkpoint gives a 0.0039 GAP boost over the best single checkpoint that we evaluated, as shown in Table \ref{table:single_model}.

\subsubsection{Ensemble Single Models into One Graph}
\ \\

As the competition requires the final model to be in a single tensorflow graph, it is not viable to ensemble single models' outputs. Instead, we build \emph{ensemble graph} at the evaluation/inference stage, and overwrite the (untrained) float16 weights tensor with previously trained float32 single model weights tensors.\\

\noindent
{\it Code snippet to build ensemble graph}
\small{
\begin{verbatim}
  with tf.variable_scope("tower"):
    result = model[0].create_model(
        model_input,
        ...,
        cluster_size = FLAGS.netvlad_cluster_size if \
                        type(FLAGS.netvlad_cluster_size) is int \
                        else FLAGS.netvlad_cluster_size[0],
        hidden_size = FLAGS.netvlad_hidden_size if \
                        type(FLAGS.netvlad_hidden_size) is int \
                        else FLAGS.netvlad_hidden_size[0])
    if FLAGS.ensemble_num> 1:
        predictions_lst = [result["predictions"]]
        for ensemble_idx in range(1,FLAGS.ensemble_num):
            with (tf.variable_scope("model"+str(ensemble_idx))):                                
                result2 = model[ensemble_idx].create_model(
                    model_input,
                    ...,
                    cluster_size = FLAGS.netvlad_cluster_size[ensemble_idx],
                    hidden_size = FLAGS.netvlad_hidden_size[ensemble_idx])
            predictions_lst.append(result2["predictions"])

    if FLAGS.ensemble_num==1:
      predictions = result["predictions"]
    else:
      predictions = 0
      for ensemble_idx in range(FLAGS.ensemble_num):
        predictions += predictions_lst[ensemble_idx] * 
                        FLAGS.ensemble_wts[ensemble_idx]
                        
\end{verbatim}
}
\noindent

To sum up our steps:
\begin{enumerate}
\item Train all single models in default float32 precision and average checkpoints.
\item For each potential ensemble model combination, build ensemble graph as above in float16 precision without training.
\item Populate ensemble graph's float16 weight tensors with single model's float32 trained weights.

\item Tune ensemble combinations and coefficients based on validate GAP.
\end{enumerate}

\section{Experiment}
We trained and evaluated 28 models with same architecture but different cluster size $K$ and hidden size $H$ mentioned in \ref{subsec:single_model}. We use the convention K$x$-H$y$ to denote model with cluster size $K=x$ and hidden size $H=y$. If a model has larger size but same or worse score than another model, we mark it an inferior model and eliminate it from the candidate list for later ensemble. There are 18 models left on the list, which is detailed in Table \ref{table:single_model}. The 10 eliminated models are K20-H1600, K64-H1024, K150-H600, K128-H512, K16-H1024, K32-H400, K128-H1024, K8-H1024, K32-H800 and K8-H800.

\setlength{\tabcolsep}{4pt}
\begin{table}
\begin{center}
\caption{The 18 single models, sorted by decreasing size and score. The two numbers in model name refer to the cluster size and hidden size. The letter in parenthesis is our model code name. Model sizes are in MB. The GAPs are evaluated on our validate set but adjusted by the constant delta to represent public leaderboard equivalent GAP.}
\label{table:single_model}
\begin{tabular}{cccccc}
\hline
Model         & Float32 & Float16 &Compression & Best avg  & Best single  \\
&  model size &  model size & rate & ckpt GAP & ckpt GAP\\
\hline
K24-H1440 (Y) & 714.93 & 381.40 & 46.7\% & 0.8744 & 0.8682 \\
K32-H1280 (G) & 679.73 & 358.85 & 47.2\% & 0.8731 & 0.8669 \\
K100-H800 (F) & 663.87 & 339.76 & 48.8\% & 0.8721 & 0.8677 \\
K16-H1280 (X) & 594.59 & 316.21 & 46.8\% & 0.8718 & 0.8665 \\
K32-H1024 (H) & 549.24 & 286.85 & 47.8\% & 0.8717 & 0.8665 \\
K20-H960 (K)  & 469.18 & 245.31 & 47.7\% & 0.8717 & 0.8661 \\
K16-H800 (L)  & 384.27 & 199.61 & 48.1\% & 0.8712 & 0.8666 \\
K64-H512 (J)  & 365.53 & 186.11 & 49.1\% & 0.8702 & 0.8664 \\
K100-H400 (A) & 357.21 & 180.93 & 49.3\% & 0.8685 & 0.8647 \\
K32-H600 (N)  & 339.72 & 174.20 & 48.7\% & 0.8684 & 0.8648 \\
K32-H512 (S)  & 297.27 & 151.85 & 48.9\% & 0.8683 & 0.8647 \\
K16-H512 (O)  & 263.13 & 134.71 & 48.8\% & 0.8665 & 0.8641 \\
K8-H512 (T)   & 246.07 & 126.15 & 48.7\% & 0.8660 & 0.8627 \\
K16-H400 (Q)  & 217.05 & 110.50 & 49.1\% & 0.8658 & 0.8619 \\
K10-H400 (E)  & 207.04 & 105.47 & 49.1\% & 0.8644 & 0.8626 \\
K32-H256 (R)  & 175.78 & 88.85  & 49.5\% & 0.8616 & 0.8593 \\
K10-H300 (M)  & 168.87 & 85.58  & 49.3\% & 0.8605 & 0.8589 \\
K16-H256 (P)  & 158.65 & 80.21  & 49.4\% & 0.8594 & 0.8576 \\ 
\hline
\end{tabular}
\end{center}
\end{table}
\setlength{\tabcolsep}{1.4pt}

\begin{figure}
\centering
\includegraphics[height=4.5cm]{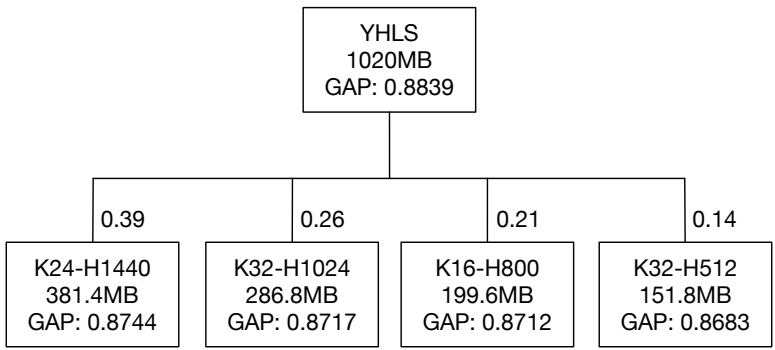}
\caption{Illustration of our best ensemble model YHLS, where each letter represents a single model. K$x$-H$y$ in each cell denotes the single model with cluster size $K=x$ and first hidden layer size $H=y$. GAP reported was evaluated on local validate set and adjusted to be Kaggle public leaderboard equivalent. Model size is reported in each cell. (0.39, 0.26, 0.21, 0.14) are the ensemble weights of each model, respectively.}
\label{fig:ensemble}
\end{figure}

We then tried out different combinations of single models, subject to the model size constraint of 1GB uncompressed. Details of ensemble models are shown in Table \ref{table:ensemble_model}. Our final selected 7th place model is the YHLS ensemble with ensemble weights = $(0.39, 0.26, 0.21, 0.14)$, see Fig.~\ref{fig:ensemble}.

\setlength{\tabcolsep}{4pt}
\begin{table}
\begin{center}
\caption{Our best ensemble models. The letters refer to the single models in Table \ref{table:single_model}. Model sizes are in MB. Public/private delta is public/private leaderboard GAP minus local GAP. The YHLS model is our 7th place model. Some LB scores are left blank because we did not do inference on test set with those models.}
\label{table:ensemble_model}
\begin{tabular}{ccccccc}
\hline
Ensemble & Model & Local & Public LB & Private LB & Public & Private  \\
& size & GAP & GAP & GAP & delta & delta\\
\hline
YHLS     & 1019.7     & 0.8821    & 0.8839       & 0.8832        & 0.0018               & 0.0011                \\
YFH      & 1008.0     & 0.8819   &               &                &                      &                       \\
GHLS     & 997.2      & 0.8818    & 0.8838       & 0.8832        & 0.0020               & 0.0014                \\
YGLP     & 1020.1     & 0.8814    &               &                &                      &                       \\
YLSQRP   & 1012.4     & 0.8811   &               &                &                      &                       \\
GHLRP    & 1015.0     & 0.8811    & 0.8832       & 0.8825        & 0.0021               & 0.0014                \\
GFX      & 1014.8     & 0.8810   &               &                &                      &                       \\
FJSOQR   & 1012.5     & 0.8810    & 0.8828       & 0.8822        & 0.0018               & 0.0012                \\
GFAB     & 1004.0     & 0.8808    & 0.8826       & 0.8821        & 0.0018               & 0.0013                \\
FJSO     & 812.4      & 0.8808    &               &                &                      &                       \\
LNSOTQP  & 977.2      & 0.8803    &               &                &                      &                       \\
GF       & 698.9      & 0.8785    &               &                &                      & \\
\hline
\end{tabular}
\end{center}
\end{table}
\setlength{\tabcolsep}{1.4pt}

\section{Conclusions}
In this paper we summarized our 7th place solution to the 2nd YouTube-8M Video Understanding Challenge. We chose the same Gated NetVLAD model architecture and trained 28 models with different hyperparameters and varying sizes. We applied three techniques to ensemble single models into a constrained-size tensorflow graph: averaging checkpoints, float16 compression, and building ensemble graph. \\ 

In our experiments, we found that in the Gated NetVLAD model, hidden size $H$ can create a bottleneck for information representation. Sacrificing cluster size $K$ in exchange for hidden size $H$ can achieve better results for constrained-size models. In ensemble, we noticed that more smaller size single models do not always beat fewer larger size single models. The optimal number of models in the ensemble is 3 to 4 in our framework. We believe this is due to the boosting effect of ensemble diminishing when we add too many models. For example, even if the ensemble of two smaller models $\alpha,\beta$ show better GAP score than a single model $\gamma$ with the same total size, after further ensembling with more models $\eta\theta$, the ensemble $\gamma\eta\theta$ may beat ensemble $\alpha\beta\eta\theta$---the boosting effect between $\alpha\beta$ and $\eta\theta$ is not as good as between $\gamma$ and $\eta\theta$ as some of the model variety boost is already ``used'' between $\alpha$ and $\beta$.\\

Had we had more time, we would explore other models and techniques since we had already done a fairly thorough search in gated NetVLAD model's hyperparameter space and tried nearly all the ensemble combinations. Given more time, we would further experiment 8-bit quantization. Should it work out, we would train models with different architectures such as LSTM, Bag-of-visual-words and Fischer Vector and fit them in the gained extra space in the model. In parallel, we could have tried distillation technique since it gives a boost in performance without needing extra model size.

\clearpage

\bibliographystyle{splncs}
\bibliography{egbib}
\end{document}